\PassOptionsToPackage{table}{xcolor}
\documentclass{article}
\usepackage{spconf,amsmath,graphicx,hyperref}
\usepackage{times}
\usepackage{latexsym}
\usepackage[T1]{fontenc}
\usepackage[utf8]{inputenc}
\usepackage{microtype}
\usepackage{inconsolata}
\usepackage{amsmath,amssymb}
\usepackage{graphicx}
\usepackage{array}
\usepackage{multirow}
\usepackage{cuted}
\usepackage{enumitem}
\usepackage{placeins}
\usepackage{booktabs}
\usepackage[table]{xcolor}
\usepackage{colortbl}
\title{PTGPO: Prefix Tree Group Policy Optimization}
\name{Ruolin Chen$^{1,2,3\star}$,
Yang Li$^{3\star}$,
Jihang Wang$^{1,2\star}$,
Zhuoya Zhao$^{4\star}$,
Yuchen Liu$^{3}$,
Jiaming Xu$^{3\dagger}$,
Qian Zhang$^{1\dagger}$
\thanks{$\star$Equal contribution. $\dagger$Corresponding author (xujiaming1@xiaomi.com, q.zhang@ia.ac.cn).}
}

\address{
$^{1}$ Brain-inspired Cognitive AI Lab, Institute of Automation,
Chinese Academy of Sciences\\
$^{2}$ School of Artificial Intelligence,
University of Chinese Academy of Sciences\\
$^{3}$ HyperAI Team, Xiaomi Corporation\\
$^{4}$ University of Ottawa
}
\begin{document}
\ninept
\maketitle
\begin{abstract}
Training language agents on long-horizon interactive tasks is challenging because sparse terminal rewards provide weak credit signals for intermediate actions. Existing group-based methods either assign a shared trajectory-level advantage or estimate step-level advantages from state groups, which may conflate the current action with subsequent decisions. We propose Prefix Tree Group Policy Optimization (PTGPO), a critic-free method that constructs prefix trees over trajectory suffixes within anchor-state groups. PTGPO assigns each tree edge a credit given by the difference between its child and parent empirical mean returns, capturing the conditional return change revealed by each additional action. These prefix-tree credits provide fine-grained learning signals without requiring a critic or additional rollouts. Experiments on interactive benchmarks ALFWorld, WebShop and Sokoban show that PTGPO consistently improves over RL baselines, demonstrating its effectiveness. 

\end{abstract}

\section{Introduction}
Recent advances~\cite{liu2025deepseek,yang2025qwen3} in large language models (LLMs) have extended their capabilities beyond static text generation toward interactive reasoning and decision-making. LLM-powered agents can perceive, reason, and act across diverse environments, including embodied tasks~\cite{shridhar2020alfworld,wang2023voyager}, web navigation~\cite{yao2022webshop,gou2025navigating}, and interactive games~\cite{SchraderSokoban2018,xu2023language}. Unlike single-turn generation, these applications require long-horizon planning and coherent sequences of actions, where intermediate decisions shape subsequent observations and task completion. Reinforcement learning (RL) provides a natural post-training framework for improving such agents through interaction and task-level feedback~\cite{qi2025webrl,jin2025search,wang2025ragen}. A central challenge is translating this feedback into effective learning signals for the individual decisions made throughout an episode.

Group-based reinforcement learning methods such as GRPO~\cite{shao2024deepseekmath} have become an effective critic-free paradigm for LLM training, where multiple responses can be sampled for the same prompt and compared using verifiable outcome rewards. These methods have demonstrated strong performance on single-turn tasks, particularly mathematical reasoning. However, extending these methods to long-horizon multi-turn agentic tasks remains challenging~\cite{wang2025ragen}. In such environments, an episode may involve many interaction steps, while rewards are often sparse, so assigning a single trajectory-level advantage to every action obscures which decisions actually caused success or failure~\cite{wang2026milestone,tan2026hindsight}. Recent methods such as GiGPO~\cite{feng2026group} mitigate this issue by grouping steps with identical environment states. However, computing step-level advantages from trajectory outcomes within such groups may conflate the grouped action with later decisions, creating a credit-assignment mismatch.

% Yet these methods still stop at grouping steps by their current and past information~\cite{wang2026milestone}, leaving open how to assign credit after the grouped step.

\begin{figure}[t]
    \centering
    \includegraphics[width=1\linewidth]{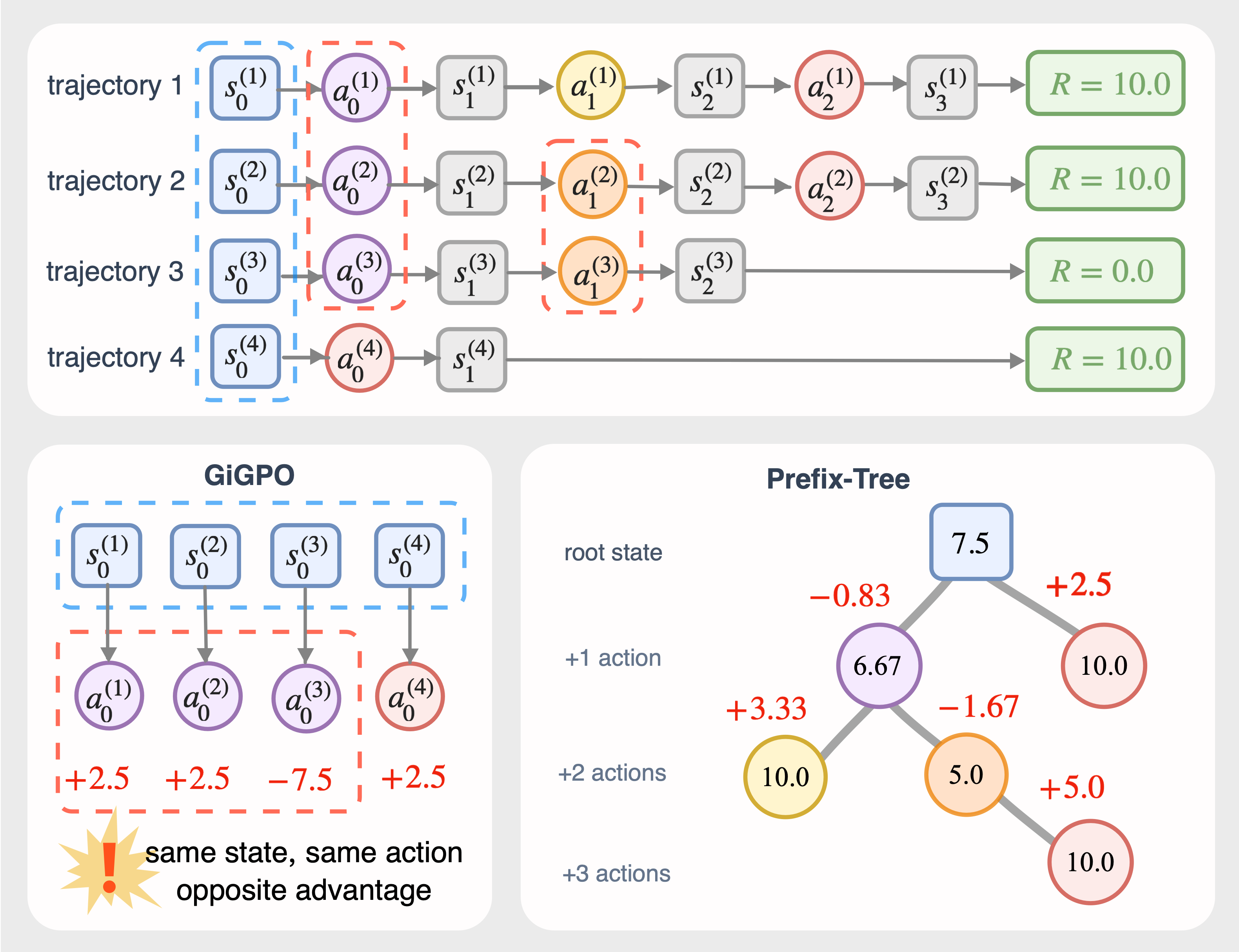}
    \caption{
Comparison of state-group and prefix-tree credit assignment under the undiscounted setting ($\gamma=1$). Identical actions may receive conflicting advantages when their subsequent actions lead to different returns. PTGPO recursively groups trajectory suffixes by action prefix, where node values denote empirical mean returns and edge labels denote child-parent mean differences. Colors indicate identical actions.
}
    \label{fig:overview}
\end{figure}

As illustrated in Fig.~\ref{fig:overview}, even when two samples share the same current state and take the same action, their terminal rewards may differ, or even have opposite signs, because their subsequent actions diverge. This difference should not be attributed to the current action, but rather to the subsequent action that actually determines the outcome. However, these decisive actions often occur after the trajectories have entered sparse or non-identical states, where reliable state-level groups can no longer be formed. The unresolved return variation should therefore be carried forward along the shared action prefix and progressively decomposed as subsequent actions are revealed. When the trajectories branch into groups with different conditional mean returns, credit can be assigned to the newly revealed branching action.

% Specifically, the return associated with action is not determined only by the local state-action pair. It also depends on the future suffix. As illustrated in Fig.~\ref{fig:overview}, even when two samples share the same current state and take the same action, they may still receive different terminal rewards. Directly using such reward to compute step-level advantages may assign different or even contradictory credits to the same local state-action pair. In such cases, attributing the return difference solely to the current action can have high variance, because the outcome may have been caused by a later decision. A faithful credit assignment mechanism should therefore refine the conditioning set along future action prefixes and assign credit only when revealing an additional action changes the conditional expected return.

\begin{figure*}[t]
    \centering
    \includegraphics[width=0.95\linewidth]{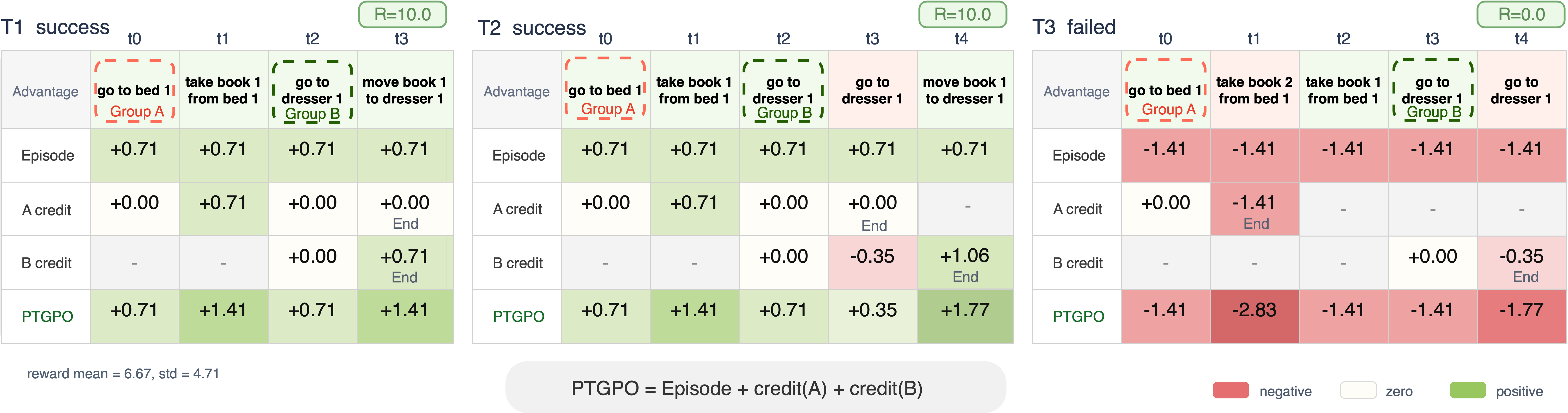}
    \caption{Illustration of PTGPO advantage decomposition under the undiscounted setting \((\gamma=1)\). Episode-level advantages are shared by all actions in the same trajectory, while prefix-tree credits redistribute return residuals to branching decisions that change the conditional mean return. ``End'' marks the termination of a prefix decomposition.}
    \label{fig:advantage_visualization}
\end{figure*}

% In this work, we propose Prefix Tree Group Policy Optimization (PTGPO), a critic-free policy optimization method that performs suffix-aware credit assignment through prefix-wise conditional mean decomposition. Given a group of sampled trajectories, PTGPO constructs a prefix tree for each root state group. Each node corresponds to a state-action prefix, and its value is estimated nonparametrically as the empirical mean return of all samples passing through that prefix. For each parent-child edge in the tree, PTGPO assigns credit equal to the difference between the child and parent empirical mean returns. This quantity measures the conditional return increment introduced by the newly revealed action. Across three challenging agentic benchmarks ALFWorld, Webshop and Sokoban, PTGPO consistently outperforms existing prompting and RL baselines. Our analysis further shows that the proposed prefix-tree credit assignment provides effective step-level learning signals, making PTGPO a practical solution for long-horizon agentic tasks with sparse rewards.

In this work, we propose Prefix Tree Group Policy Optimization (PTGPO), a critic-free method for fine-grained credit assignment in long-horizon tasks. PTGPO constructs a prefix tree for each anchor-state group, recursively grouping trajectory suffixes by shared action prefixes. For each parent-child edge in the tree, PTGPO assigns credit equal to the difference between the child and parent empirical mean returns. This quantity measures the conditional return increment introduced by the newly revealed action. Across three challenging agentic benchmarks ALFWorld, WebShop and Sokoban, PTGPO consistently outperforms existing prompting and RL baselines. Our analysis further shows that the proposed prefix-tree credit assignment provides effective step-level learning signals, making PTGPO a practical solution for long-horizon agentic tasks with sparse rewards.

\section{Method}
For each sampled trajectory $i$, let its length be $T_i$. For each sampled step $(i,t)$, we define the discounted return as
\begin{equation}
\hat R^{(i)}_t
=
\sum_{u=t}^{T_i}
\gamma^{u-t} r^{(i)}_u,
\end{equation}
where $\gamma\in[0,1]$ is the discount factor. In sparse terminal-reward tasks, intermediate rewards are zero.

Following GiGPO, we use the same anchor-state matching strategy to construct step-level root groups over sampled rollouts. Let $\tilde s$ denote an anchor state, and let
\begin{equation}
\mathcal G(\tilde s)
=
\{(i,t)\mid s_t^{(i)}=\tilde s\}
\end{equation}
be the set of all step occurrences that share this anchor state.

For each group, PTGPO builds a prefix tree. Let $\mathbf a_d$ be a sequence of $d$ actions. The corresponding node is the set
\begin{equation}
p(\tilde s,\mathbf a_d)=\bigl\{(i,t)\in\mathcal G(\tilde s)\;\big|\;(a_t^{(i)},\ldots,a_{t+d-1}^{(i)})=\mathbf a_d\bigr\}.
\end{equation}
Samples with the same action sequence of length $d$ are included. The root is $p(\tilde s,\varnothing)=\mathcal G(\tilde s)$. If $\mathbf a_{d+1}$ extends $\mathbf a_d$ by one action $a_d$,
then the node $p(\tilde s,\mathbf a_{d+1})$ is a child of $p(\tilde s,\mathbf a_d)$, and we have
\begin{equation}
p(\tilde s,\mathbf a_{d+1})
\subseteq
p(\tilde s,\mathbf a_d).
\end{equation}
Thus, starting from the root, PTGPO recursively splits each node according to the next action, merging sampled steps with the same future action prefix into the same branch.

% The root is $p(\tilde s,\varnothing)=\mathcal G(\tilde s)$. Starting from the root, PTGPO follows its remaining action sequence, merging steps with the same next action into a branch and recursively repeating this merging within each branch for subsequent actions.

For any nonempty node, its empirical mean return is
\begin{equation}
\hat\mu\bigl(p(\tilde s,\mathbf a_d)\bigr)
=
\frac{
1
}{
|p(\tilde s,\mathbf a_d)|
} \displaystyle\sum_{(i,t)\in p(\tilde s,\mathbf a_d)}
\hat R_t^{(i)}.
\end{equation}

% For a sampled step $(i,t)$ with $s_t^{(i)}=\tilde{s}$, we abbreviate the node corresponding to its length-$d$ future action prefix $\mathbf a_d$ as $p_d=p(\tilde{s},\mathbf a_d)$, with $p_0=p(\tilde{s},\varnothing)$. 
For a sampled step $(i,t)$ with $s_t^{(i)}=\tilde{s}$,
consider its prefix-tree path, where
$\mathbf a_d=(a_t^{(i)},\ldots,a_{t+d-1}^{(i)})$
and $\mathbf a_0=\varnothing$. The unnormalized state-group step advantage~\cite{feng2026group} then admits the following decomposition along a path expanded to depth $D$:
\begin{equation}
\begin{aligned}
\hat R_t^{(i)}-\hat\mu\bigl(p(\tilde{s},\varnothing)\bigr)
&=
\sum_{d=0}^{D-1}
\Delta\bigl(
p(\tilde{s},\mathbf a_d),
p(\tilde{s},\mathbf a_{d+1})
\bigr) \\
&\quad +
\left[
\hat R_t^{(i)}-\hat\mu\bigl(p(\tilde{s},\mathbf a_D)\bigr)
\right],
\end{aligned}
\end{equation}
% \begin{equation}
% \hat R_t^{(i)}-\hat\mu(p_0)=\sum_{d=0}^{D-1}\Delta(p_d,p_{d+1})+\left[\hat R_t^{(i)}-\hat\mu(p_D)\right],
% \end{equation}
where
\begin{equation}
\Delta\bigl(
p(\tilde{s},\mathbf a_d),
p(\tilde{s},\mathbf a_{d+1})
\bigr)
=
\hat\mu\bigl(p(\tilde{s},\mathbf a_{d+1})\bigr)
-
\hat\mu\bigl(p(\tilde{s},\mathbf a_d)\bigr)
\end{equation}
% \begin{equation}
% \Delta(p_d,p_{d+1})=\hat\mu(p_{d+1})-\hat\mu(p_d)
% \end{equation}
is the credit assigned to the newly added action $a_{t+d}^{(i)}$ for each $(i,t)\in p(\tilde{s},\mathbf a_{d+1})$, reflecting the change in mean return from parent to child, while differences due to later actions remain within $p(\tilde{s},\mathbf a_{d+1})$ for further decomposition. A node can be expanded only if $|p(\tilde{s},\mathbf a_d)|\ge\tau_{\mathrm{expand}}$.

\begin{table*}[t]
\centering
\caption{Performance on ALFWorld and WebShop. We train each RL method with 3 random seeds and report the mean and standard deviation. The best performances are highlighted in bold.}
\label{tab:main_results}
\resizebox{\textwidth}{!}{
\begin{tabular}{llccccccccc}
\toprule
\multirow{2}{*}{\textbf{Type}} 
& \multirow{2}{*}{\textbf{Method}}
& \multicolumn{7}{c}{\textbf{ALFWorld}}
& \multicolumn{2}{c}{\textbf{WebShop}} \\
\cmidrule(lr){3-9} \cmidrule(lr){10-11}
& 
& \textbf{Pick}
& \textbf{Look}
& \textbf{Clean}
& \textbf{Heat}
& \textbf{Cool}
& \textbf{Pick2}
& \textbf{All}
& \textbf{Score}
& \textbf{Succ.} \\
\midrule
\rowcolor{gray!10}
\multicolumn{11}{l}{\textit{Closed-Source Model}} \\

Prompting
& GPT-4o
& 75.3 & 60.8 & 31.2 & 56.7 & 21.6 & 49.8 & 48.0 & 31.8 & 23.7\\

Prompting
& Gemini-2.5-Pro
& 92.8 & 63.3 & 62.1 & 69.0 & 26.6 & 58.7 & 60.3 & 42.5 & 35.9 \\

\midrule
\rowcolor{gray!10}
\multicolumn{11}{l}{\textit{Qwen2.5-1.5B-Instruct}} \\

Prompting
& Direct Prompt
& 5.9 & 5.5 & 3.3 & 9.7 & 4.2 & 0.0 & 4.1 & 23.1 & 5.2\\

Prompting
& ReAct
& 17.4 & 20.5 & 15.7 & 6.2 & 7.7 & 2.0 & 12.8 & 40.1 & 11.3\\

Prompting
& Reflexion
& 35.3 & 22.2 & 21.7 & 13.6 & 19.4 & 3.7 & 21.8 & 55.8 & 21.9 \\

RL Training
& PPO
& 64.8$_{\scriptscriptstyle \pm 3.5}$ & 40.5$_{\scriptscriptstyle \pm 6.9}$ & 57.1$_{\scriptscriptstyle \pm 4.9}$ & 60.6$_{\scriptscriptstyle \pm 6.6}$ & 46.4$_{\scriptscriptstyle \pm 4.0}$ & 47.4$_{\scriptscriptstyle \pm 1.9}$ & 54.4$_{\scriptscriptstyle \pm 3.1}$
& 73.8$_{\scriptscriptstyle \pm 3.0}$ & 51.5$_{\scriptscriptstyle \pm 2.9}$\\

RL Training
& RLOO
& 88.3$_{\scriptscriptstyle \pm 3.0}$ & 52.8$_{\scriptscriptstyle \pm 8.6}$ & 71.0$_{\scriptscriptstyle \pm 5.9}$ & 62.8$_{\scriptscriptstyle \pm 8.7}$ & 66.4$_{\scriptscriptstyle \pm 5.5}$ & 56.9$_{\scriptscriptstyle \pm 4.7}$ & 69.7$_{\scriptscriptstyle \pm 2.5}$
& 73.9$_{\scriptscriptstyle \pm 5.6}$ & 52.1$_{\scriptscriptstyle \pm 6.7}$ \\

RL Training
& GRPO
& 85.3$_{\scriptscriptstyle \pm 1.5}$
& 53.7$_{\scriptscriptstyle \pm 8.0}$
& 84.5$_{\scriptscriptstyle \pm 6.8}$
& 78.2$_{\scriptscriptstyle \pm 7.9}$
& 59.7$_{\scriptscriptstyle \pm 5.0}$
& 53.5$_{\scriptscriptstyle \pm 5.6}$
& 72.8$_{\scriptscriptstyle \pm 3.6}$
& 75.8$_{\scriptscriptstyle \pm 3.5}$
& 56.8$_{\scriptscriptstyle \pm 3.8}$ \\

RL Training
& GiGPO
& 93.6$_{\scriptscriptstyle \pm 6.7}$
& 82.1$_{\scriptscriptstyle \pm 8.0}$
& 90.3$_{\scriptscriptstyle \pm 6.9}$
& 89.7$_{\scriptscriptstyle \pm 5.5}$
& 77.5$_{\scriptscriptstyle \pm 4.0}$
& 73.3$_{\scriptscriptstyle \pm 11.4}$
& 84.8$_{\scriptscriptstyle \pm 5.5}$
& 81.6$_{\scriptscriptstyle \pm 4.6}$
& 67.5$_{\scriptscriptstyle \pm 4.8}$ \\

% RL Training
% & HGPO
% & 93.5$_{\scriptscriptstyle \pm 4.0}$
% & 83.0$_{\scriptscriptstyle \pm 9.0}$
% & 92.7$_{\scriptscriptstyle \pm 6.4}$
% & 96.1$_{\scriptscriptstyle \pm 6.8}$
% & \textbf{86.7}$_{\scriptscriptstyle \pm 6.1}$
% & 86.7$_{\scriptscriptstyle \pm 7.9}$
% & 90.2$_{\scriptscriptstyle \pm 2.2}$
% & 85.6$_{\scriptscriptstyle \pm 2.9}$
% & 71.5$_{\scriptscriptstyle \pm 4.0}$ \\

\rowcolor{blue!10}
\textbf{RL Training}
& \textbf{PTGPO (Ours)}
& \textbf{96.6}$_{\scriptscriptstyle \pm 3.6}$ & \textbf{87.6}$_{\scriptscriptstyle \pm 6.9}$ & \textbf{97.4}$_{\scriptscriptstyle \pm 2.3}$ & \textbf{100.0}$_{\scriptscriptstyle \pm 0.0}$ & \textbf{86.0}$_{\scriptscriptstyle \pm 6.6}$ & \textbf{89.3}$_{\scriptscriptstyle \pm 8.2}$ & \textbf{92.6}$_{\scriptscriptstyle \pm 2.5}$
& \textbf{87.1}$_{\scriptscriptstyle \pm 0.4}$ & \textbf{74.3}$_{\scriptscriptstyle \pm 1.5}$ \\

\midrule
\rowcolor{gray!10}
\multicolumn{11}{l}{\textit{Qwen2.5-7B-Instruct}} \\

Prompting
& Direct Prompt
& 33.4 & 21.6 & 19.3 & 6.9 & 2.8 & 3.2 & 14.8 & 26.4 & 7.8\\

Prompting
& ReAct
& 48.5 & 35.4 & 34.3 & 13.2 & 18.2 & 17.6 & 31.2 & 46.2 & 19.5\\

Prompting
& Reflexion
& 62.0 & 41.6 & 44.9 & 30.9 & 36.3 & 23.8 & 42.7 & 58.1 & 28.8\\

RL Training
& PPO
& 92.3$_{\scriptscriptstyle \pm 4.0}$ & 64.0$_{\scriptscriptstyle \pm 8.4}$ & 92.5$_{\scriptscriptstyle \pm 2.4}$ & 89.5$_{\scriptscriptstyle \pm 7.0}$ & 80.3$_{\scriptscriptstyle \pm 2.0}$ & 68.8$_{\scriptscriptstyle \pm 8.3}$ & 80.4$_{\scriptscriptstyle \pm 2.7}$
& 81.4$_{\scriptscriptstyle \pm 3.1}$ & 68.7$_{\scriptscriptstyle \pm 5.1}$ \\

RL Training
& RLOO
& 87.6$_{\scriptscriptstyle \pm 4.3}$ & 78.2$_{\scriptscriptstyle \pm 8.3}$ & 87.3$_{\scriptscriptstyle \pm 5.8}$ & 81.3$_{\scriptscriptstyle \pm 7.6}$ & 71.9$_{\scriptscriptstyle \pm 5.2}$ & 48.9$_{\scriptscriptstyle \pm 8.4}$ & 75.5$_{\scriptscriptstyle \pm 4.6}$
& 80.3$_{\scriptscriptstyle \pm 3.2}$ & 65.7$_{\scriptscriptstyle \pm 4.0}$ \\

RL Training
& GRPO
& 90.8$_{\scriptscriptstyle \pm 5.1}$
& 66.1$_{\scriptscriptstyle \pm 6.7}$
& 89.3$_{\scriptscriptstyle \pm 5.4}$
& 74.7$_{\scriptscriptstyle \pm 6.9}$
& 72.5$_{\scriptscriptstyle \pm 5.4}$
& 64.7$_{\scriptscriptstyle \pm 7.3}$
& 77.6$_{\scriptscriptstyle \pm 5.2}$
& 79.3$_{\scriptscriptstyle \pm 2.8}$
& 66.1$_{\scriptscriptstyle \pm 3.7}$ \\

RL Training
& GiGPO
& 95.2$_{\scriptscriptstyle \pm 5.9}$
& 86.7$_{\scriptscriptstyle \pm 4.3}$
& 96.4$_{\scriptscriptstyle \pm 3.4}$
& 93.7$_{\scriptscriptstyle \pm 5.9}$
& 84.1$_{\scriptscriptstyle \pm 5.9}$
& 92.8$_{\scriptscriptstyle \pm 2.4}$
& 91.9$_{\scriptscriptstyle \pm 1.6}$ 
& 82.5$_{\scriptscriptstyle \pm 1.3}$
& 70.5$_{\scriptscriptstyle \pm 1.1}$\\

% RL Training
% & HGPO
% & \textbf{100.0}$_{\scriptscriptstyle \pm 0.0}$
% & 91.2$_{\scriptscriptstyle \pm 2.0}$
% & \textbf{97.6}$_{\scriptscriptstyle \pm 2.1}$
% & 92.4$_{\scriptscriptstyle \pm 6.6}$
% & 84.4$_{\scriptscriptstyle \pm 1.5}$
% & 87.5$_{\scriptscriptstyle \pm 7.6}$
% & 92.6$_{\scriptscriptstyle \pm 0.8}$  
% & 89.0$_{\scriptscriptstyle \pm 1.0}$
% & 78.5$_{\scriptscriptstyle \pm 1.4}$ \\

\rowcolor{blue!10}
\textbf{RL Training}
& \textbf{PTGPO (Ours)}
& \textbf{100.0}$_{\scriptscriptstyle \pm 0.0}$ & \textbf{96.0}$_{\scriptscriptstyle \pm 3.5}$ & \textbf{97.3}$_{\scriptscriptstyle \pm 2.4}$ & \textbf{97.6}$_{\scriptscriptstyle \pm 4.1}$
& \textbf{86.9}$_{\scriptscriptstyle \pm 5.0}$ & \textbf{97.2}$_{\scriptscriptstyle \pm 2.4}$ & \textbf{96.0}$_{\scriptscriptstyle \pm 1.1}$
& \textbf{90.7}$_{\scriptscriptstyle \pm 2.0}$ & \textbf{82.0}$_{\scriptscriptstyle \pm 1.6}$ \\
\bottomrule
\end{tabular}
}
\end{table*}

\begin{table}[t]
\centering
\caption{Performance on Sokoban. We train each RL method with 3 random seeds and report the mean and standard deviation. The best performances are highlighted in bold.}
\label{tab:sokoban}
\begin{tabular}{ccc}
\toprule
\textbf{Type} & \textbf{Method} & \textbf{Sokoban [6$\times$6]} \\
\midrule
Prompting   & Direct Prompt & 11.7  \\
RL Training & GRPO       & 73.9$_{\scriptscriptstyle \pm 1.6}$ \\
RL Training & GiGPO      & 77.6$_{\scriptscriptstyle \pm 2.5}$ \\
% RL Training & HGPO      & 76.6$_{\scriptscriptstyle \pm 1.6}$ \\
\rowcolor{blue!10}
\textbf{RL Training} & \textbf{PTGPO} &  \textbf{85.2}$_{\scriptscriptstyle \pm 1.6}$ \\
\bottomrule
\end{tabular}
\end{table}

% Each nonempty child receives its incoming edge credit, but can be expanded further only if its own count meets the threshold. Unresolved variation remains within nodes where expansion stops.

PTGPO normalizes each credit by the standard deviation of returns in its root group:
\begin{equation}
\hat\sigma\bigl(p(\tilde{s},\varnothing)\bigr)
=
\operatorname{Std}
\left(
\left\{
\hat R_t^{(i)}
\mid
(i,t)\in p(\tilde{s},\varnothing)
\right\}
\right).
\end{equation}
% \begin{equation}
% \hat\sigma(p_0)
% =
% \operatorname{Std}
% \left(\{\hat R_t^{(i)}\mid(i,t)\in p_0\}\right).
% \end{equation}
For action $a_k^{(i)}$, PTGPO sums the normalized edge credits from starting steps $t=0,\ldots,k$:
\begin{equation}
A_k^{\mathrm{PT},(i)}
=
\sum_{t=0}^{k}
\frac{
\Delta\Bigl(
p\bigl(s_t^{(i)},\mathbf a_{k-t}\bigr),
p\bigl(s_t^{(i)},\mathbf a_{k-t+1}\bigr)
\Bigr)
}{
\hat\sigma\bigl(p(s_t^{(i)},\varnothing)\bigr)
}.
\end{equation}
Each edge credit is normalized by the standard deviation of the root group associated with its starting step $t$, with each term capturing a distinct conditional-return innovation under a different anchor-conditioned prefix decomposition. In practice, the accumulated magnitude remains well controlled because the maximum expansion depth of each prefix tree is limited and valid root groups become increasingly sparse along the trajectory.

% \begin{equation}
% A_k^{\mathrm{PT},(i)}=\sum_{t=0}^{k}\frac{1}{\hat\sigma(p_0)}\Delta(p_{k-t},p_{k-t+1}).
% \end{equation}
% In the summand indexed by $t$, $p_{k-t}$ and $p_{k-t+1}$ are consecutive nodes in the prefix tree rooted at $p_0=\mathcal G(s_t^{(i)})$. Accordingly, each edge credit is normalized by the standard deviation of the root group associated with its starting step $t$, with each term capturing a distinct conditional-return innovation under a different anchor-conditioned prefix decomposition. In practice, the accumulated magnitude remains well controlled because the maximum expansion depth of each prefix tree is limited and valid root groups become increasingly sparse along the trajectory.

Finally, PTGPO is combined with the base episode advantage:
\begin{equation}
A^{\mathrm{episode},(i)}
=
\frac{
\hat R^{(i)}_{T_i}-\mu_R
}{
\sigma_R
},
\end{equation}
where $\mu_R$ and $\sigma_R$ are the mean and standard deviation of terminal rewards across the group.
\begin{equation}
A^{(i)}_k
=
A^{\mathrm{episode},(i)}
+
 A^{\mathrm{PT},(i)}_k.
\end{equation}
Here $A^{\mathrm{episode},(i)}$ is sequence-level and shared by all actions in the response, while $A^{\mathrm{PT},(i)}_k$ provides step-level credit by decomposing the suffix-dependent return along the prefix tree. Fig.~\ref{fig:advantage_visualization} gives an intuitive example of the resulting advantage decomposition. PTGPO shifts credit from uniformly rewarding or penalizing every step to emphasizing the key decision points that separate successful and failed trajectories.

Then the clipped policy optimization objective of PTGPO is:

\begin{equation}
\begin{aligned}
\mathcal J(\theta)
=
\mathbb E
\Big[
\min \Big(
&\rho^{(i)}_k(\theta)A^{(i)}_k, \\
&\operatorname{clip}
\big(\rho^{(i)}_k(\theta),1-\epsilon,1+\epsilon\big)
A^{(i)}_k
\Big)
\Big].
\end{aligned}
\end{equation}
where 
\begin{equation}
\rho^{(i)}_k(\theta)
=
\frac{
\pi_\theta(a^{(i)}_k\mid s^{(i)}_k)
}{
\pi_{\theta_{\mathrm{old}}}(a^{(i)}_k\mid s^{(i)}_k)
}.\end{equation}

% \begin{table}[t]
% \centering
% \caption{Time cost (s) at epoch 160 on ALFWorld.}
% \label{tab:time_cost}
% \small
% \begin{tabular}{c|cccc}
% \toprule
% \textbf{Epoch} & \textbf{Common} & \textbf{GRPO} & \textbf{GiGPO} & \textbf{PTGPO}\\
% \midrule
% 160   & 210.968 & 0.192 & 0.559 & 1.713 \\
% \bottomrule
% \end{tabular}
% \end{table}

\section{Experiments}

\subsection{Setup}
\textbf{Benchmarks.} We train agents on three challenging multi-turn benchmarks: ALFWorld~\cite{shridhar2020alfworld}, WebShop~\cite{yao2022webshop} and Sokoban~\cite{SchraderSokoban2018}. ALFWorld is a text-based embodied environment where agents complete household tasks through multi-step interaction. WebShop is a web-based interactive benchmark where LLM agents navigate a simulated shopping website to purchase items matching user instructions. Sokoban is a visual grid-based game benchmark where agents move boxes onto target locations through multi-step actions.

\noindent
\textbf{Baselines.} We compare PTGPO against baselines across paradigms: (1) Closed-source models: GPT-4o~\cite{hurst2024gpt} and Gemini-2.5-Pro~\cite{comanici2025gemini}. (2) Prompting methods: ReAct~\cite{yao2022react} and Reflexion~\cite{shinn2023reflexion}. (3) RL training methods: PPO~\cite{schulman2017proximal}, RLOO~\cite{ahmadian2024back}, GRPO~\cite{shao2024deepseekmath} and GiGPO~\cite{feng2026group}, covering both critic-based and group-based policy optimization methods. For VLMs, we focus on comparisons with group-based RL methods.

\noindent
\textbf{Implementation.} We use Qwen2.5-1.5B-Instruct and Qwen2.5-7B-Instruct~\cite{qwen25} as the base models for ALFWorld and WebShop, and Qwen2.5-VL-3B-Instruct~\cite{qwen25vl} for Sokoban. For fair comparison, all RL methods use identical training configurations. We train each method for 160 training iterations with a discount factor of $\gamma=0.95$ and a learning rate of $1\times10^{-6}$. In each iteration, we sample 16 training instances and generate a group of 8 trajectories for each instance, resulting in 128 rollouts per iteration. Each episode allows up to 50 environment steps in ALFWorld and up to 15 environment steps in WebShop and Sokoban. We use 8 NVIDIA H20 96GB GPUs for training.

\begin{table}[t]
\centering
\caption{Ablation study of PTGPO on ALFWorld and WebShop using Qwen2.5-1.5B-Instruct, and on Sokoban using Qwen2.5-VL-3B-Instruct. The best performances are highlighted in bold.}

\label{tab:ptgpo_ablation}
\setlength{\tabcolsep}{3pt}
\begin{tabular}{lccc}
\toprule
\textbf{Variant} & \textbf{ALFWorld} & \textbf{WebShop} & \textbf{Sokoban} \\
\midrule
\textbf{PTGPO} & \textbf{92.6}$_{\scriptscriptstyle \pm 2.5}$
& 74.3$_{\scriptscriptstyle \pm 1.5}$ &\textbf{85.2}$_{\scriptscriptstyle \pm 1.6}$\\ 
\midrule
\rowcolor{gray!10}
\multicolumn{4}{l}{\textit{Advantage Composition}} \\
\quad w/o Episode-level Adv & 92.4$_{\scriptscriptstyle \pm 1.1}$ &  71.8$_{\scriptscriptstyle \pm 3.9}$ &  
81.5$_{\scriptscriptstyle \pm 4.4}$ \\
\quad w/o Prefix-tree Adv & 72.8$_{\scriptscriptstyle \pm 3.6}$ &  56.8$_{\scriptscriptstyle \pm 3.8}$& 73.9$_{\scriptscriptstyle \pm 1.6}$  \\
\midrule
\rowcolor{gray!10}
\multicolumn{4}{l}{\textit{Aggregation Strategy}} \\
\quad Mean & 84.3$_{\scriptscriptstyle \pm 1.6}$ &  71.4$_{\scriptscriptstyle \pm 2.1}$ &  82.3$_{\scriptscriptstyle \pm 2.3}$ \\
\quad Nearest Anchor & 91.5$_{\scriptscriptstyle \pm 2.3}$ &  73.2$_{\scriptscriptstyle \pm 1.6}$& 84.1$_{\scriptscriptstyle \pm 1.0}$  \\
\midrule
\rowcolor{gray!10}
\multicolumn{4}{l}{\textit{Prefix Depth}} \\
% \quad $d=1$ & 84.8$_{\scriptscriptstyle \pm 5.5}$
% & 67.5$_{\scriptscriptstyle \pm 4.8}$ & 77.6$_{\scriptscriptstyle \pm 2.5}$\\
\quad $d=2$ & 87.6$_{\scriptscriptstyle \pm 4.4}$ & 70.0$_{\scriptscriptstyle \pm 0.9}$  & 78.6$_{\scriptscriptstyle \pm 5.1}$ \\
\quad $d=4$ & 88.3$_{\scriptscriptstyle \pm 2.2}$ & 72.9$_{\scriptscriptstyle \pm 1.0}$ & 83.3$_{\scriptscriptstyle \pm 6.6}$ \\
\quad $d=8$ & 92.1$_{\scriptscriptstyle \pm 1.2}$ & \textbf{74.5}$_{\scriptscriptstyle \pm 1.9}$ & 83.6$_{\scriptscriptstyle \pm 1.6}$\\
\midrule
\rowcolor{gray!10}
\multicolumn{4}{l}{\textit{Expansion Threshold}} \\
\quad ${\tau_{\mathrm{expand}}}=4$ & 90.0$_{\scriptscriptstyle \pm 3.6}$&
72.0$_{\scriptscriptstyle \pm 3.7}$& 83.6$_{\scriptscriptstyle \pm 3.4}$\\
\quad ${\tau_{\mathrm{expand}}}=6$ & 85.5$_{\scriptscriptstyle \pm 1.1}$&
71.1$_{\scriptscriptstyle \pm 2.2}$& 81.5$_{\scriptscriptstyle \pm 1.2}$\\
\quad ${\tau_{\mathrm{expand}}}=8$ &84.3$_{\scriptscriptstyle \pm 2.9}$ &73.3$_{\scriptscriptstyle \pm 3.8}$& 81.8$_{\scriptscriptstyle \pm 1.7}$\\
\bottomrule
\end{tabular}
\end{table}

\subsection{Experiment Results}

\subsubsection{Main Results} Table~\ref{tab:main_results} shows that PTGPO consistently outperforms prompting methods and RL baselines across multi-turn agentic tasks. On Qwen2.5-1.5B, PTGPO achieves 92.6 on ALFWorld and 74.3\% success on WebShop, improving over GRPO by 19.8 and 17.5 points, respectively. With Qwen2.5-7B, PTGPO further improves to 96.0 on ALFWorld and 82.0\% success rate on WebShop. On the visual Sokoban benchmark, PTGPO reaches 85.2 as shown in Table~\ref{tab:sokoban}, outperforming GRPO and GiGPO, demonstrating its effectiveness for both LLM and VLM agents. To further examine the training dynamics, Figure~\ref{fig:training_curve}(a) reports the episode success rate on ALFWorld as a representative example. PTGPO reaches around 0.8 success rate after about 75 steps, while GiGPO reaches a similar level after about 105 steps and GRPO fails to do so within 160 steps, showing better training efficiency.

\subsubsection{Ablation Studies}
Table~\ref{tab:ptgpo_ablation} evaluates the main components and design choices of PTGPO. Unless otherwise specified, the full method uses unrestricted prefix-tree expansion with $\tau_{\mathrm{expand}}=2$ and sums all valid prefix credits associated with each action. Removing the prefix-tree advantage reduces performance by 19.8, 17.5, and 11.3 points on ALFWorld, WebShop, and Sokoban, respectively, confirming that fine-grained prefix credit is the primary source of improvement. In comparison, removing the episode-level advantage has little effect on ALFWorld but causes larger drops on WebShop and Sokoban, indicating that the trajectory-level signal provides complementary global supervision.

We next compare different strategies for combining the credits assigned to the same action by multiple anchor states. Mean aggregation consistently underperforms the default summation strategy, with a particularly large drop on ALFWorld. Using only the nearest valid anchor performs substantially better than mean aggregation, but remains below credit summation on all three benchmarks. These results suggest that different anchor states provide complementary credit signals: restricting learning to the nearest anchor discards useful longer-range evidence, whereas averaging can dilute informative signals when only a small number of valid anchors are available. Summation preserves and accumulates these signals. Moreover, because an action can only receive credit from trees that reach it, the number of accumulated terms is bounded by the tree depth and is further limited by sparse state matching and the expansion threshold.

\begin{figure}[t]
    \centering
    \includegraphics[width=1\linewidth]{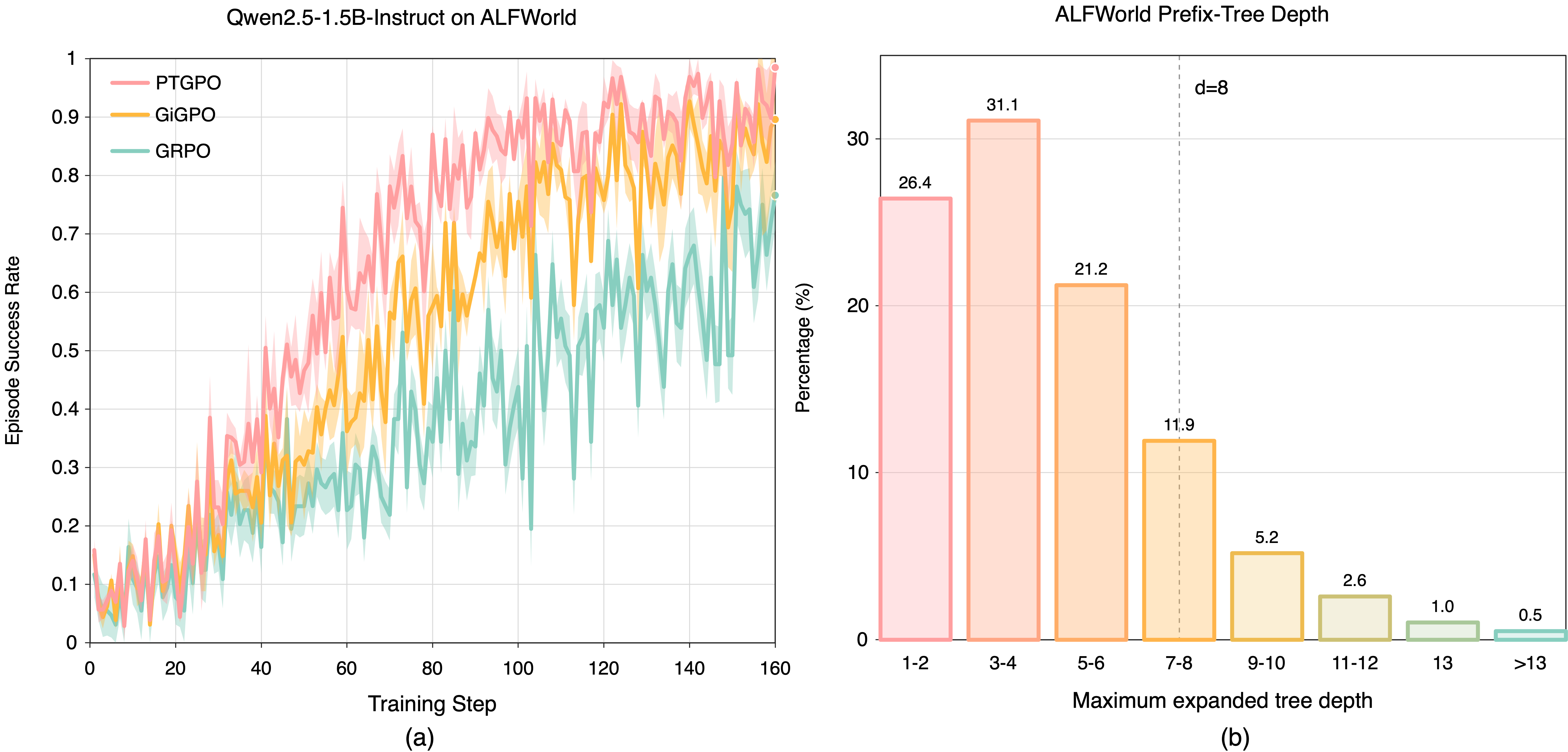}
    \caption{(a) Episode Success Rate during Training on ALFWorld. (b) Distribution of maximum expanded tree depth at epoch 160 on ALFWorld.}
    \label{fig:training_curve}
\end{figure}

Increasing the maximum prefix depth generally improves performance. Expanding the depth from $2$ to $8$ improves ALFWorld from 87.6 to 92.1 and WebShop from 70.0 to 74.5, while Sokoban improves from 78.6 to 83.6. Depth $8$ is already close to unrestricted expansion, consistent with Fig.~\ref{fig:training_curve}(b), which shows that most trees terminate within eight levels. Finally, increasing $\tau_{\mathrm{expand}}$ generally degrades performance. Although a larger threshold yields more reliable node-level return estimates, it prevents many sparsely supported branches from being expanded and consequently removes useful fine-grained credits. Overall, $\tau_{\mathrm{expand}}=2$ provides the best balance between estimation reliability and credit coverage.
% \textbf{Ablation Studies. }Table~\ref{tab:ptgpo_ablation} verifies the contribution of each component in PTGPO. Unless otherwise specified, PTGPO uses full prefix-tree expansion with $\tau_{\mathrm{expand}}=2$, and each ablation changes one factor at a time. Removing the prefix-tree advantage causes large drops on all benchmarks, while removing the episode-level advantage leads to smaller degradation. Increasing the prefix depth generally improves performance. Taking ALFWorld as an example, Figure~\ref{fig:training_curve}(b) shows that most prefix trees are expanded within depth 8, which explains why $d=8$ already achieves performance close to full PTGPO. Varying $\tau_{\mathrm{expand}}$ reflects the trade-off between reliable mean estimation and fine-grained credit assignment: larger thresholds require more samples for expansion, but may stop the tree earlier and lose useful prefix-level distinctions. The default value $\tau_{\mathrm{expand}}=2$ achieves the best overall performance in our experiments.

\subsubsection{Computational Overhead}
We measure the method-specific advantage computation cost on ALFWorld with Qwen2.5-1.5B-Instruct at epoch 160. PTGPO shares the same rollout, probability computation, and policy update pipeline as other group-based methods, with a common cost of about 210.968s per iteration, and introduces additional computation only for prefix-tree construction and prefix-wise advantage estimation. The method-specific advantage computation costs are 0.192s for GRPO, 0.559s for GiGPO, and 1.713s for PTGPO. Despite the additional overhead introduced by prefix-tree construction, PTGPO's extra computation accounts for only 0.812\% of the common training cost and has a negligible impact on overall training efficiency.

\section{Conclusion}

In this paper, we propose PTGPO, a novel reinforcement learning method for long-horizon interactive agents with sparse rewards. PTGPO mitigates the credit assignment problem by decomposing trajectory returns through prefix-wise conditional mean increments, assigning step-level credit to actions according to how future action prefixes change expected returns. Experiments on interactive benchmarks show that PTGPO consistently improves over existing RL baselines. Further analyses demonstrate the importance of prefix-tree advantages and show that PTGPO introduces only minor computational overhead without requiring extra rollouts or critic models.

\bibliographystyle{IEEEbib}
\bibliography{strings,refs}

\end{document}